\documentclass[11pt, a4paper]{article}

\usepackage[margin=1in]{geometry}
\usepackage[utf8]{inputenc}
\usepackage[T1]{fontenc}
\usepackage{amsmath}
\usepackage{hyperref}
\usepackage{authblk} 
\usepackage{multirow}
\usepackage{adjustbox}
\usepackage{color,soul}
\usepackage[table,xcdraw]{xcolor}

\title{Metric-Guided Synthetic Image Data Rendering for Deep Learning compatible with Agentic AI}

\author[1,2,3,+]{Martina Radoynova}
\author[1,2,+]{Samuel Pantze}
\author[1,2]{Trina De}
\author[2,*]{Ulrik Günther}
\author[1,2,3,4,*]{Artur Yakimovich}
\affil[1]{Center for Advanced Systems Understanding (CASUS), Görlitz, Germany}
\affil[2]{Helmholtz-Zentrum Dresden-Rossendorf e. V. (HZDR), Dresden, Germany}
\affil[3]{Institute of Computer Science, University of Wrocław, Wrocław, Poland}
\affil[4]{Cluster of Excellence Physics of Life, TU Dresden, Dresden, Germany}
\affil[*]{correspondence: a.yakimovich@hzdr.de, ulrik.guenther@hzdr.de}
\affil[+]{these authors contributed equally to this work} 

\begin{document}
\maketitle

\begin{abstract}
Deep learning computer vision for scientific applications requires collecting and annotating large datasets in a laborious, expensive and error-prone process. Synthetic data generation through 3D modelling and rendering may simplify this process and increase the accuracy of annotations by generating them programmatically. However, minimising the domain gap between real and synthetic images visually is subjective and lacks systematic quantitative guidance. We present GraNatPy, a Python package with metrics to guide improvement of the rendered scene. We show that quantifiable increase in realism, diversity and size of rendered dataset correlates with improved visual perception of the scene and higher zero-shot performance of an object detection model. Furthermore, we demonstrated using photographs of virological plaque assays that gradient similarity affects performance on small object detection, which can be improved by mixing real and synthetic data. Finally, we turn procedural data rendering into an agentic skill (SynthClaw) to automate the procedural parameter optimisation.
\end{abstract}

\section*{Introduction}
Deep learning\cite{lecun2015deep} models are revolutionising biomedical research by providing tools that were unimaginable a decade ago. These range from protein folding and biomolecular interaction prediction\cite{abramson2024accurate} to single-cell analysis enabled through rapid high-fidelity cell segmentation\cite{stringer2021cellpose, Stardist18}. However, these outstanding examples were driven by data-rich, well-formulated problems. At the same time, the vast majority of biomedical data is scarce or too sensitive to use. Together with other computational approaches (reviewed in \cite{yakimovich2021labels}), synthetic data lends itself to alleviating this barrier.

The practicality of computational synthesis of training data exists in two planes. Firstly, this approach to model training allows for virtually infinite amounts of data. Secondly, it features an impeccable and consistent ground truth, which is unattainable through human labelling\cite{warfield2004simultaneous}. Furthermore, even simple rule-based techniques like domain randomisation\cite{gupta2016synthetic,tremblay2018training} have proven highly effective in boosting generalisation capabilities of deep learning algorithms and now constitute a staple of the deep-learning toolset.

Most approaches to date are based either on simulating the real data\cite{dosovitskiy2017carla, wood2021fake, denninger2023blenderproc2} or generating it using generative AI approaches \cite{shrivastava2017learning, mueller2018ganerated, zhang2021datasetgan} (reviewed in\cite{nikolenko2021synthetic, lu2023machine}). A notable example of the former is the procedural synthetic data simulation (rendering) in autonomous driving\cite{dosovitskiy2017carla, bibi2024synthetic}. While some report exceptional zero-shot performance, e.g. face recognition using synthetic data alone\cite{wood2021fake}, the effectiveness of both simulation and generation-based approaches suffers from the domain gap \cite{tobin2017domain} when it comes to real-world performance. It is because of the very existence of the domain gap that some approaches explore domain-independent pretraining, proposing an entirely different avenue\cite{nakamura2024scaling}. However, the vast majority of synthetic data approaches simply aim to narrow the domain gap by increasing size, realism and the diversity of the synthetic datasets\cite{gaidon2016virtual, tobin2017domain, ros2016synthia, shrivastava2017learning}. Some researchers report that prioritising diversity alongside realism achieves optimal performance\cite{Staniszewskietal2025}. Yet, while the size of the dataset is easy to measure, both diversity and realism are not, hampering principled synthetic data generation.

In this work, we explored procedural synthetic data rendering for object detection in digital photographs of a biological assay (VACVPlaque\cite{de2025digital} dataset of virological plaque assay\cite{yakimovich2019high}) using the open-source 3D software Blender\cite{blender}. To minimise the domain gap in a guided and principled manner, we investigated quantitative metrics for both realism and diversity. For the latter, we use entropy\cite{shannon1948communication}. For the former, we explored both the Fréchet Inception Distance (FID), the Learned Perceptual Image Patch Similarity (LPIPS)\cite{zhang2018unreasonable}, as well as the gradient similarity through the image Naturalness Factor ($dN_f$), based on normalised gradients\cite{abgaryan2025regularized, gong2014image} (see Methods). We evaluated the performance through training on the synthetic dataset and testing on the real dataset strategy\cite{gupta2016synthetic}, which we further refer to as zero-shot. Through this evaluation, we demonstrated that using FID, $dN_f$ and entropy as guidance allowed us to reach optimal zero-shot performance on object detection in this biological assay. To turn these metrics into a reusable capability, we developed \href{https://github.com/casus/GraNatPy}{GraNatPy} package which includes a Python version for $N_f$ and $dN_f$ and diversity metrics calculation we contributed. Finally, to lower the entry barrier for procedural synthetic data rendering, we formulated the headless interaction with Blender as an open-source agentic skill for the OpenClaw platform, available free and open-source on ClawHub.

\section*{Results}
The conventional deep learning pipeline requires careful collection and annotation of the training data for model training. Should the model performance remain unsatisfactory, training data is often identified as the culprit, prompting further data collection efforts (Fig. 1a). To alleviate this limitation, we designed a synthetic data generation pipeline, allowing to generate virtually limitless amounts of new data (Fig. 1b). As a case study we chose the instance segmentation task, known for its laborious image annotation requirements. Specifically, we focused on virological plaque assay\cite{dulbecco1952production} segmentation using the VACVPlaque open dataset\cite{de2025digital}. In this assay, cultured cells are infected with a virus of interest and incubated until distinctive lesions appear. Then the cells are fixed, stained, washed and photographed. To tackle instance segmentation in the photographs, we used a single-shot architecture derived from StarDist\cite{Stardist18} with a custom loss described in\cite{de2025single}.

To mimic the assay photographs, we constructed a custom Blender scene (see Methods) by modelling a six-well plate and creating procedurally generated objects imitating stained monolayers of adhesion cultured cells (wells) and circular lesions (plaques) in it. For this, we used Blender's Geometry Nodes and Shader Nodes systems. These are function block-based interfaces for creating and manipulating geometry and materials programmatically. In addition, we created a Blender-internal post-processing pipeline to further randomise the rendered images and export segmentation masks. We then devised a command-line interface-based approach, where the scene's procedural parameters can be controlled using a script. This approach allows us to batch-generate 2D images with high similarity to the real photographs accompanied by nearly perfect ground truth (GT). With dataset size limitation removed, we focused on two other factors that have been reported to influence model performance\cite{Staniszewskietal2025} through the domain gap: synthetic data closeness to the real data (here referred to as realism) and the dataset diversity.

\begin{figure}[ht]
 \centering
  \includegraphics[width=0.8\textwidth]{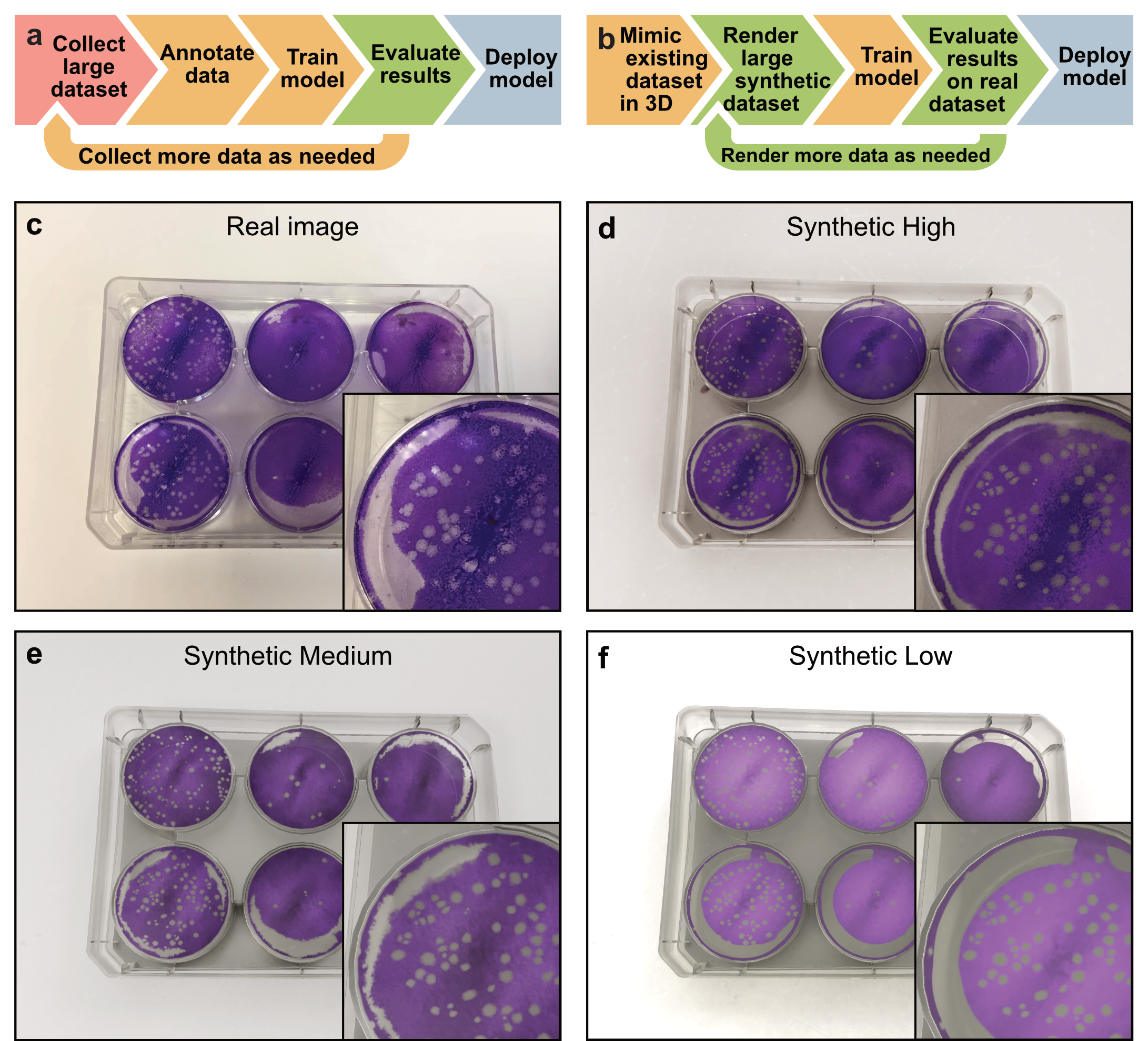}
\caption{\label{fig:fig1} \textbf{3D rendered synthetic data generation pipeline and results compared to conventional data collection approach.} (a) conventional data collection pipeline. (b) 3D rendering synthetic data pipeline. Here, green, orange and red denote the relatively low, medium and high effort levels, respectively, associated with a particular step. (c) depicts a real image taken from a virological plaque assay plate. (d-f) depict 3D rendered replica of the image in (a), rendered in high, medium and low realism, respectively.}
\end{figure}

\subsection*{Selecting a Metric for Synthetic Data Realism}
To test whether we can visually improve synthetic image realism and find a suitable quantitative measure of realism, we have constructed a scene rendering of an image nearly identical to a specific data point in our real dataset (Fig. 1c,d, further referred to as high realism). To ensure that the metric of our choice is capable of capturing what we perceive visually, we rendered the same image in two grades of lower realism (Fig. 1e,f, medium and low, respectively). Next, with the synthetic images and the ground truth (GT) at hand, we measured the mean square error (MSE), normalised root mean square error (NRMSE), structured similarity index measure (SSIM)\cite{ssim}, peak signal-to-noise ratio (PSNR)\cite{PSNR}, learned perceptual image patch similarity (LPIPS)\cite{zhang2018unreasonable}, and the Fréchet Inception Distance (FID)\cite{fid}. To be able to calculate and compare FIDs, we cut the images into identically-sized patches of 6 evenly spaced sub-images. Additionally, we measured image naturalness\cite{gong2014image} ($N_f$), which effectively represents a measure of normalised gradients and their closeness to natural scene photographs\cite{gong2014image,abgaryan2025regularized}. To ensure that this metric is not biased towards exclusively natural scenes, we measured the difference in image naturalness between the GT and the test image ($dN_f$, see Methods) (Tab. \ref{tab:metrics-realism}). Results suggest that LPIPS, FID and $dN_f$ scale best with the perceived realism from low through medium to high. To prioritise the most commonly used metric, we selected FID for further evaluation as a metric of dataset realism.

\begin{table}[ht]
\centering
\begin{tabular}{lccccccc}
\hline
Perceived Realism & MSE$\downarrow$ & NRMSE$\downarrow$ & SSIM$\uparrow$ & PSNR$\uparrow$ & LPIPS$\downarrow$  &$dN_f\downarrow$ & FID$\downarrow$\\ \hline
High & 0.0102 & 0.1434 & 0.7808 & 19.90 dB & \textbf{0.339}  &\textbf{0.046} & \textbf{122.80} \\
Medium & \textbf{0.0101} & \textbf{0.1426} & \textbf{0.7973} & \textbf{19.95 dB} & 0.343  &0.266 & 148.53 \\
Low & 0.0295 & 0.2435 & 0.7728 & 15.30 dB & 0.431  & 0.386 & 166.65 \\
\hline
\end{tabular}
\caption{Measured and perceived realism of our rendered images. Here, MSE is mean square error, NRMSE is normalised root mean square error, SSIM is the structural similarity index measure, PSNR is peak signal-to-noise ratio, LPIPS is the learned perceptual image patch similarity, $dN_f$ is the difference of the synthetic $N_f$ to the real $N_f$, and FID is the Fréchet inception distance. \textbf{Closest to real}.}
\label{tab:metrics-realism}
\end{table}

\subsection*{Selecting a Metric for Synthetic Data Diversity}
Next, we explored suitable metrics to evaluate diversity in our synthetic datasets (see Tab. \ref{tab:metrics-diversity}). Our goal was to investigate how the metrics would scale with image variability in the dataset depending on realism and size. For this, first we generated the synthetic image datasets in three levels of realism (similar to Tab. 1 and Fig. 1c-f) and two different sizes, 100 (small, S) images and 1000 images (large, L) by randomising the procedural parameters as described in Methods. Then we constructed a control dataset with no variability in pixel values (No Change) by copying the same image over multiple times, aimed to be compared to the synthetic and real datasets.  Finally, we measured dataset-wide brightness variability and Shannon's entropy ($H$)\cite{shannon1948communication} across all three groups of images (see Methods).

\begin{table}[ht]
\centering
\begin{tabular}{lcccccc}
\hline
Dataset & Brightness Variability ($\uparrow$) & Entropy ($H\uparrow$) \\ \hline
High-S & \textbf{700.07} & \textbf{5.41} &  \\
Medium-S & 459.73 & 4.71  \\
Low-S & 622.15 & 3.73  \\
\hline
High-L & \textbf{691.36} & \textbf{6.05} \\
Medium-L & 472.42 & 5.13 \\
Low-L & 606.43 & 4.21 \\
\hline
\textit{No Change} & \textit{0.00} & \textit{0.00} \\
\hline
\textit{Real} & \textit{1240.33} & \textit{5.82} \\
\hline
\end{tabular}
\caption{Quantitative assessment of image diversity across synthetic datasets, a real dataset, and the \textit{No~Change} control dataset. \textit{Brightness Variability} denotes the mean pixel-wise variance of grayscale intensities calculated across all images within each dataset. \textit{Entropy} represents the mean pixel-wise Shannon entropy calculated from the intensity distributions at corresponding spatial locations. \textbf{Highest values} indicate greater variability within the respective metric.}
\label{tab:metrics-diversity}
\end{table}

Both metrics were slightly higher for the larger datasets compared to their smaller equivalents. Both brightness variability and $H$ of the highest perceived realism sets had the highest scores measuring diversity and were closest to the real image dataset values. However, only entropy scaled well with perceived realism (from low through medium to high), suggesting co-dependency. While such scaling was not necessarily expected from a diversity metric, given the nature of the entropy, it seemed the most appropriate diversity metric for our purposes, which motivated our choice.

To ensure that both realism and diversity metrics are easily accessible in a Pythonic environment, we created the \href{https://github.com/casus/GraNatPy}{GraNatPy} package. In this package, we implemented brightness variability and entropy and re-implemented image naturalness in Python. Additionally, we included existing Python metrics like FID as dependencies.

\begin{figure}[ht]
    \centering
    \includegraphics[width=0.8\linewidth]{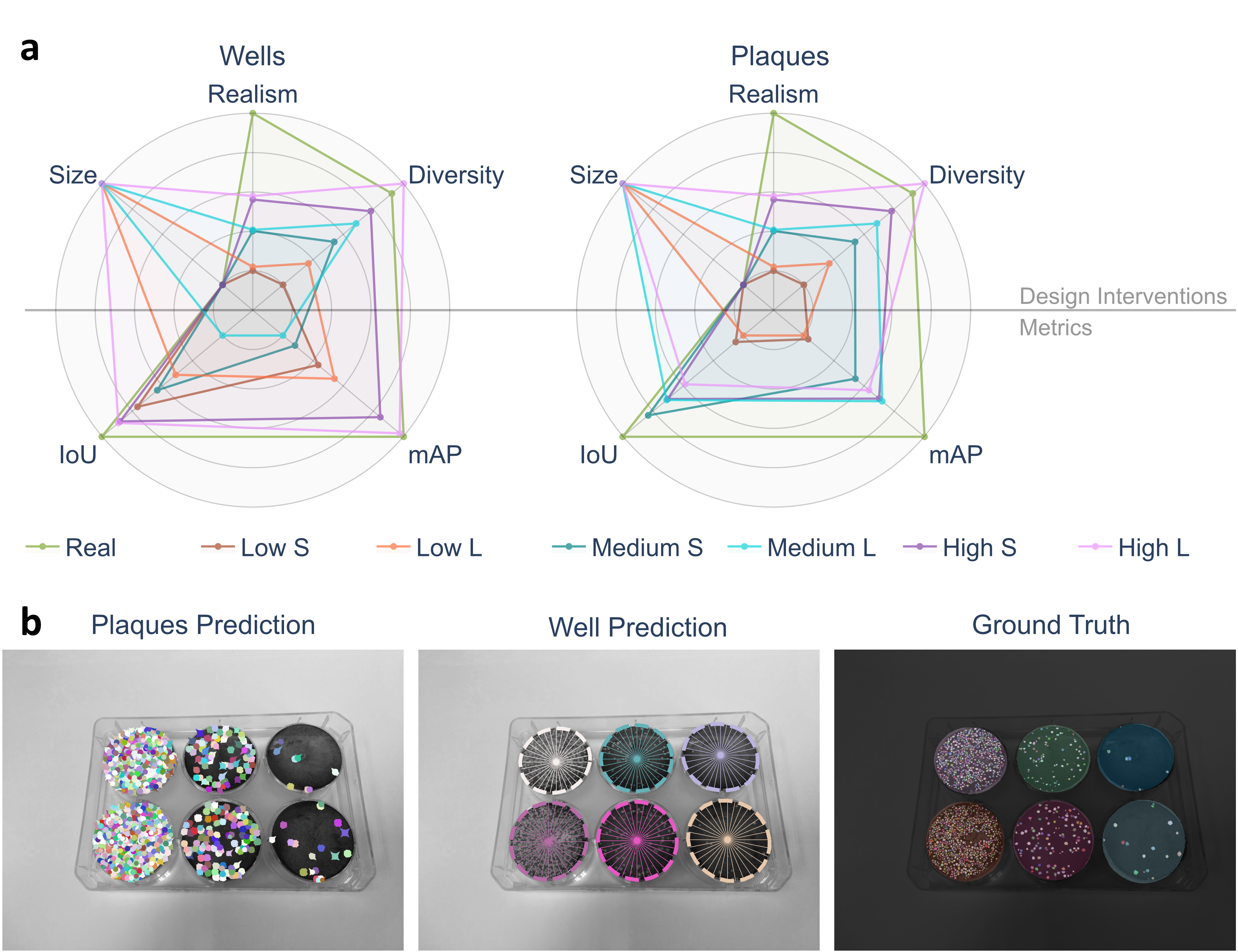}
    \caption{\textbf{a} Performance of models trained on Real and Synthetic images, measured with design interventions (naturalness, diversity and size) and segmentation metrics (IoU and mAP). For values, see Tab. \ref{tab:metrics_wells}. \textbf{b} Example predictions of plaque and well segmentation from a model trained on High-L.}
    \label{fig:fig2}
\end{figure}

\subsection*{Zero-shot Performance Comparison}
With quantitative metrics for size, realism and diversity at hand, we explored the relationship between these factors and our synthetic data suitability for model pretraining. While several strategies for such evaluation can be selected, for simplicity, we decided to assess the suitability of the synthetic datasets through zero-shot model performance. Namely, through training models on synthetic data and evaluating on real human-annotated data.

Instance segmentation annotations in the virological plaque assay dataset \cite{de2025digital} exist at two levels. First, the large objects are the wells of the multi-titre assay plate (here, a 6-well plate). Second, the small objects are the lesions caused by virus propagation through the monolayer of cultured cells. While the model we selected\cite{de2025single} for the evaluation is performing the instance segmentation task in a single shot, results are shown in Tab. \ref{tab:metrics_wells} and Tab. \ref{tab:metrics_plaques} separately for the ease of comparison. Performance was evaluated using Intersection over Union ($IoU_R$)\cite{Stardist18, de2025single} and mean average precision (mAP)\cite{Stardist18, de2025single}. Additionally, to present the findings across the multiple conditions in an intuitive manner, we collated them as two radar diagrams for large (wells) and small (plaques) objects, respectively (Fig. \ref{fig:fig2}a).

\begin{table}[ht!]
\begin{adjustbox}{width=\textwidth}
\centering
\setlength{\tabcolsep}{10pt}
\begin{tabular}{c|ccc|cc}
\hline
\multirow{2}{*}{Dataset Name}
& \multicolumn{3}{c|}{Dataset Properties}
& \multicolumn{2}{c}{Instance Segmentation Metrics} \\
\cline{2-6}
& Size & Realism (FID$\downarrow$) & Diversity ($H\uparrow$) & $IoU_R\uparrow$ & mAP$\uparrow$ \\
\hline
High-S & 100 & \underline{107.74 } & \underline{5.4135} & \underline{0.8721 (± 0.0125)}  &    \underline{0.9173 (± 0.0346)}  \\
Medium-S & 100 & 146.95 & 4.7101 & 0.7272 (± 0.0727)  &    0.2884 (± 0.0171) \\
Low-S & 100 & 196.53 & 3.7254 &  0.8067 (± 0.0835)  &    0.4361 (± 0.0212) \\
\hline
High-L & 1000 & \textbf{103.27} & \textbf{6.0471} & \textbf{ 0.8842 (± 0.0027)} & \textbf{0.9558 (± 0.0095) }\\
Medium-L & 1000 & 145.50 & 5.1312 & 0.4638 (± 0.0163)  &    0.2118 (± 0.0025) \\
Low-L & 1000 & 191.64  & 4.2150 & 0.6530 (± 0.0446)  &    0.5405 (± 0.0665) \\
\hline
\textit{Mix} & \textit{90+10} & \textit{94.01} & \textit{5.5136} & \textit{\textbf{0.9289 (± 0.0022)}} &   \textit{\textbf{0.9438 (± 0.0421)}}\\
\hline
\textit{Real} & \textit{100} & \textit{0} & \textit{5.8189} & \textit{0.9507 (± 0.0048)} &\textit{0.9820 (± 0.0143)} \\
\hline
\end{tabular}
\end{adjustbox}
\caption{\textbf{Zero-shot segmentation of wells (large objects)}.  Performance on real test holdout of synthetic datasets with varying properties: \textit{Size} - S(mall) and L(arge), \textit{Realism} - measured with FID and \textit{Diversity} - measured with Shannon  Entropy; and evaluated on Instance Segmentation Metrics: \textit{Intersection over Union over recalled objects ($IoU_R$)} and \textit{Average Precision (mAP)} at threshold \(\tau=0.5\), compared to model trained on Real data. \textit{Mix} refers to a combination of 10\% real and 90\% generated data. \textbf{Best performance.} \underline{Second best.}}
\label{tab:metrics_wells}
\end{table}

Performance on wells (larger objects) segmentation (see Tab. \ref{tab:metrics_wells}, Fig. \ref{fig:fig2}a) suggested that, consistent with the existing literature\cite{Staniszewskietal2025}, maximising the size, realism and diversity of the dataset leads to better zero-shot performance on the real data. Surprisingly, in the case of smaller objects (plaques), the best and second-best zero-shot performances were split across the medium-realism datasets and the small high-realism dataset (Tab. \ref{tab:metrics_plaques}, Fig. \ref{fig:fig2}b). These results may likely be explained by the fact that while being small, the plaques are numerous across both small and large datasets. This potentially alleviates the dependency on the dataset size, allowing a small, high-realism dataset to perform on par with the best-performing dataset. However, this does not explain why the S and L medium realism datasets scored higher on $IoU_R$ and mAP, respectively. 

To explore whether this observation can be connected to the similarity of the image gradients, we additionally measured the $dN_f$ of these datasets (Tab. \ref{tab:metrics_plaques}) by calculating the averaged Naturalness factor $N_f$ for each dataset and then comparing it to the averaged $N_f$ of the real dataset. Remarkably, the $dN_f$ of the Medium-S and Medium-L datasets turned out to be lower than the $dN_f$ of High-S and High-L, suggesting that the gradients in the medium realism image datasets are closer to the real image dataset than the high realism. This correlated with the unexpectedly high performance of the medium datasets on instance segmentation of small objects.

\begin{table}[h]
\centering
\begin{adjustbox}{width=\textwidth}
\setlength{\tabcolsep}{8pt}
\begin{tabular}{c|cccc|cc}
\hline
\multirow{2}{*}{Dataset Name}
& \multicolumn{4}{c|}{Dataset Properties}
& \multicolumn{2}{c}{Instance Segmentation Metrics} \\
\cline{2-7}
& Size & Realism (FID $\downarrow$) & $dNf \downarrow$ & Diversity ($H\uparrow$) & $IoU_R$ $\uparrow$ & mAP $\uparrow$ \\
\hline
High-S & 100 & \underline{107.74} & 0.3680 & \underline{5.4135} & \underline{0.2536 (± 0.0110)}  &    \underline{0.2585(± 0.0116)}  \\
Medium-S & 100 & 146.95 & \underline{0.3358} & 4.7101 & \textbf{0.2687 (± 0.0081)}  &    0.1922 (± 0.0173) \\
Low-S & 100 & 196.53 & 0.6919 & 3.7254 & 0.0618 (± 0.0096)  &    0.0686 (± 0.0144) \\
\hline
High-L & 1000 & \textbf{103.27} & 0.3913 & \textbf{6.0471} & 0.1811 (± 0.0083) & 0.2282 (± 0.0081) \\
Medium-L & 1000 & 145.50 & \textbf{0.3345} & 5.1312 & 0.2253 (± 0.0075)  &    \textbf{0.2625 (± 0.0034)} \\
Low-L & 1000 & 191.64 & 0.6662 & 4.2150 & 0.0435 (± 0.0084)  &     0.0570 (± 0.0092) \\
\hline
\textit{Mix} & \textit{90+10} & \textit{94.01} & \textit{0.3380} & \textit{5.5136} & \textit{\textbf{0.2811 (± 0.0131)}} &   \textit{\textbf{0.3280 (± 0.0133)}}\\
\hline
\textit{Real} & \textit{100} & \textit{0} & \textit{0} & \textit{5.8189}  & \textit{ 0.3290 (± 0.0332)} & \textit{0.3736 (± 0.0356)} \\
\hline
\end{tabular}
\end{adjustbox}
\caption{\textbf{Zero-shot segmentation of plaques (small objects)}. Performance on real test holdout of synthetic datasets with ranging properties: \textit{Size} - S(mall) and L(arge), \textit{Realism} - measured with $dN_f$ and \textit{Diversity} - measured with Shannon  Entropy; and evaluated on Instance Segmentation Metrics: \textit{Intersection over Union over recalled objects ($IoU_R$)} and \textit{Average Precision (mAP)} at threshold \(\tau=0.5\), compared to model trained on Real data. \textit{Mix} refers to a combination of 10\% real and 90\% generated data. \textbf{Best performance.} \underline{Second best.}}
\label{tab:metrics_plaques}
\end{table}

To test whether the instance segmentation performance on small objects (plaques) can be improved by adding real images, we mixed 90 images from the High-S dataset with 10 real images. Such a composition of the dataset slightly increased diversity and the similarity of gradients (decreased $dN_f$), while substantially increasing our measure of realism (lower FID, see Mix row in Tabs. \ref{tab:metrics_wells} and \ref{tab:metrics_plaques}). Remarkably, such an approach greatly improved the instance segmentation results on both small and large objects with only 10 real images (Tabs. \ref{tab:metrics_wells} and \ref{tab:metrics_plaques}). Notably, the small Mix dataset outperformed all large synthetic datasets, underlying the significance of dataset realism and diversity over size.  Furthermore, while the diversity of the Mix dataset increased slightly compared to High-S, it remained much lower than High-L, suggesting that realism is more important than size and diversity. Interestingly, $dN_f$ of the Mix dataset was closer (albeit slightly higher) to Medium-S than High-S, suggesting that gradient similarity may play a role in performance. However, low $dN_f$ \textit{per se} does not seem to be sufficient to achieve such exceptionally high performance (Medium-L vs. Mix). Crucially, these results demonstrate that a significant improvement in dataset realism and model performance can be achieved by adding a very small amount of real data into a larger synthetic dataset.

\newpage
\begin{figure}[ht]
    \centering
    \includegraphics[width=0.95\linewidth]{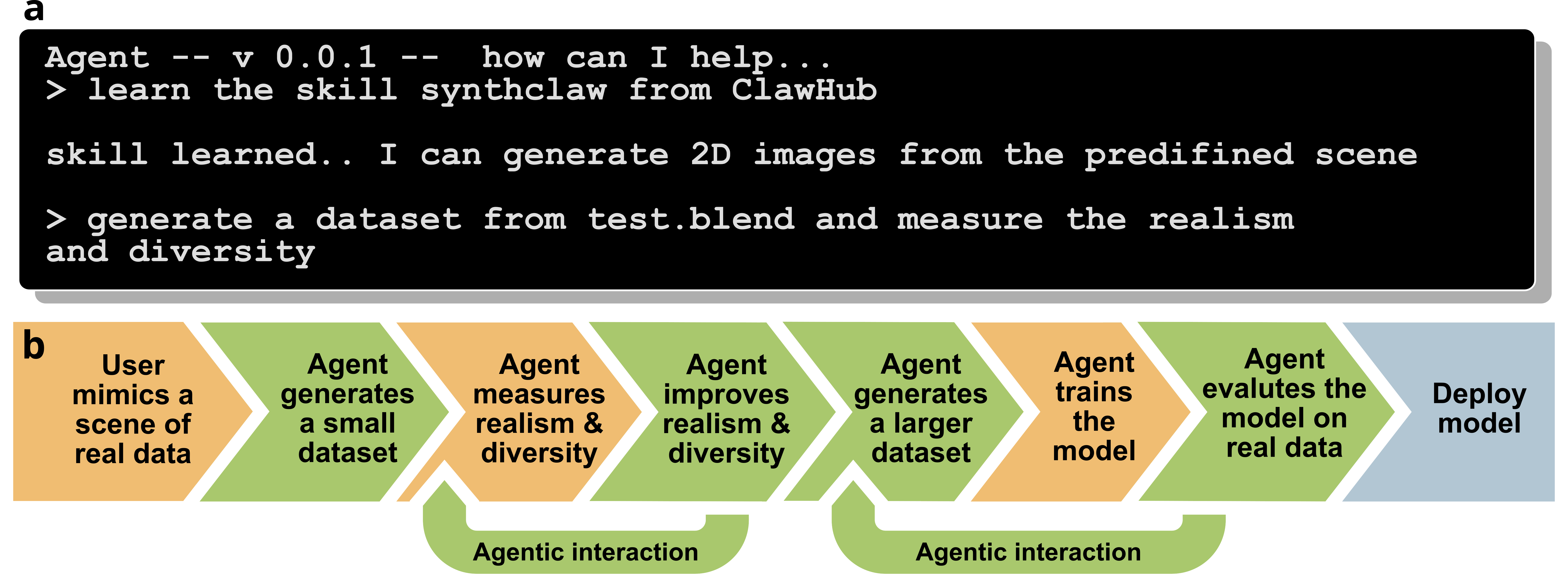}
    \caption{Metric-guided synthetic image data rendering compatible with Agentic AI. \textbf{a} A mock-up of a natural language interaction example with an agent. \textbf{b} Agentic 3D rendering synthetic data pipeline. Green and orange denote low and medium task complexity.}
    \label{fig:agent-workflow}
\end{figure}

\subsection*{Agentic Skill for Procedural Rendering}
3D modelling software has become increasingly more approachable through open-source licensing. However, using it to generate synthetic image data for scientific deep learning still requires a steep learning curve and remains inaccessible to most researchers. To lower this barrier, we developed an agentic skill allowing us to automate some of the synthesis operations through a natural language interface (Fig.~\ref{fig:agent-workflow}a). Such a skill can be incorporated into an agentic harness system (e.g. OpenClaw), which in turn uses a chat interface to execute actions, allowing it to generate images from a pre-designed scene file, modify procedural parameters exposed in the file, choose rendering quality and obtain metric measurements (Fig.~\ref{fig:agent-workflow}b). While such an approach still requires the scene to be designed by the user, it significantly simplifies further iterations, allowing the use of metrics across the synthetic dataset as a guide for an agentic workflow. 

Furthermore, the full scope of the agent's capabilities includes single and batch scene generation, realism and diversity metrics evaluation (through our \href{https://github.com/casus/GraNatPy}{GraNatPy} package), size evaluation, and model training and evaluation. Notably, while complex modifications like architectural changes to the node graph, high-level decisions about the graph structure and scene lighting and setup require user intervention, these capabilities can already be used to create a simple agentic workflow (Fig. 3b). To facilitate further development, we made the skill compatible with the \href{https://agentskills.io/home}{agentskills.io} skill standard and open-sourced it as part of this repository \href{https://github.com/casus/SynthClaw}{github.com/casus/SynthClaw}. The skill is available to install into a compatible agent harness under \href{https://clawhub.ai/ayakimovich/synthclaw}{clawhub.ai/ayakimovich/synthclaw} and licensed under the permissive MIT license.

\section*{Discussion}
Deep learning has demonstrated an incredible capacity in detecting complex patterns across scientific domains\cite{novelAI24,volpe2026roadmap}. While the list of largely solved tasks appears to be growing, the scarcity of large annotated datasets is hampering its application to niche tasks. Several approaches beyond supervised learning have been proposed\cite{yakimovich2021labels}. However, in domains like vision and tasks like instance segmentation supervised techniques are still dominating.

In this work, we looked at instance segmentation in biomedical image analysis -- a problem with one of the most cumbersome ground truth annotations. This task requires a user to delineate hundreds or thousands of individual objects at pixel precision in hundreds or thousands of data points (images) just to start the training. Furthermore, we selected a niche but important task of virological plaque assay. Introduced in the 1950s\cite{dulbecco1952production}, the plaque assay is fundamental to Virology. Today, thousands of plaque assays are performed globally, yet only one open dataset with instance segmentation level annotations exists to date\cite{de2025digital}. To tackle this gap, we developed a procedural pipeline for realistic annotated synthetic data generation using the Blender open-source rendering software\cite{blender}. We made the resulting dataset and the source files openly available\cite{rodare}.

While previous efforts of Blender-based procedural synthetic annotated datasets exist\cite{denninger2023blenderproc2}, this is the first such attempt for virological plaque assay. Beyond the plaque assay, many fundamental cell biological assays\cite{franken2006clonogenic} follow the same pattern: they are performed in a multi-titre plate and can be quantified manually or using computer-vision based instance segmentation. These assays, however, lack the annotated data for model development. It is tempting to speculate that procedural synthetic data generation employing our open-sourced Blender files may facilitate the deep-learning-based automated quantification in these cases.

Furthermore, in this work we set out to guide our synthetic data generation using a suite of quantitative metrics, which we assembled into an open source Python package GraNatPy\cite{zenodo_granatpy}. Using these metrics, we demonstrate in a set of computational experiments that  size, realism and diversity drive the best zero-shot performance of the synthetic data, corroborating the existing research\cite{Staniszewskietal2025}. However, we also demonstrate that the relationship of these factors is not always equal. When it comes to numerous small objects in an image, this equilibrium shifts. Our results show that in such cases, the size of the datasets may no longer play such an important role. Furthermore, factors like similarity of image gradients (measured by $dN_f$ here) may play a role. Finally, an increase in realism may overrule all the other factors. We also show that such an increase in the dataset realism can be effectively achieved by mixing a small (10\%) amount of real images into a synthetic dataset. 

These results also underline that the scarcity of large annotated datasets must not present the ultimate hurdle and can be overcome by procedural synthetic data generation. Acknowledging that the latter might still represent a technical hurdle for the life science community, we constructed SynthClaw\cite{zenodo_synthclaw} -- an agentic skill simplifying procedural data generation from a Blender scene using a natural language interface. While the agentic capabilities of interacting with 3D scene rendering software are an active area of development, to our knowledge, this represents the first effort on agentic procedural synthetic data rendering. While there is ongoing work by the Blender community to introduce agentic capabilities into the graphic user interface (GUI) through the Model Context Protocol (MCP, see \href{https://modelcontextprotocol.io}{modelcontextprotocol.io}), SynthClaw is aimed at headless execution and bulk dataset generation guided by the GraNatPy metrics. Finally, we argue that despite being tested on a specific task of instance segmentation in virological plaque assay, the fundamentals of this work are generalisable to other tasks. Together, our work suggests a novel path for niche life science tasks' inclusion into the ongoing deep learning revolution in science.

\section*{Acknowledgements}
This work was partially funded by the Center for Advanced Systems Understanding (CASUS) which is financed by Germany’s Federal Ministry of Research, Technology and Space (BMFTR) and by the Saxon Ministry for Science, Culture, and Tourism (SMWK) with tax funds based on the budget approved by the Saxon State Parliament. AY is supported by the Helmholtz Association Initiative and Networking Fund in the frame of Helmholtz AI as well as by the Helmholtz Foundation Model Initiative within the project “PROFOUND”.

\section*{Code availability}\label{code-availability}
Model code has been published on GitHub: \href{https://github.com/casus/HSD-WBR-Synth}{github.com/casus/HSD-WBR-Synth} and is available on Zenodo\cite{zenodo_hsd_wbr_syn}. \\
Metrics code including Naturalness calculation, dataset entropy and other metrics is available on GitHub under \href{https://github.com/casus/GraNatPy}{github.com/casus/GraNatPy}, as a  \href{https://pypi.org/project/granatpy/}{PyPI package} and on Zenodo\cite{zenodo_granatpy}.\\
The agentic skill is available on \href{https://clawhub.ai/ayakimovich/synthclaw}{ClawHub}, skill code is available on GitHub under \href{https://github.com/casus/SynthClaw}{github.com/casus/SynthClaw} and Zenodo\cite{zenodo_synthclaw}.

\section*{Data availability}
Generated datasets, as well as the 3D rendering pipelines, have been made available as a RODARE repository\cite{rodare}.

\section*{Methods}

\subsection*{Blender Procedural Pipeline}
Blender provides powerful tools for the creation of procedural datasets. In our case, at the heart of the procedural pipeline lies a \textit{Geometry Nodes} modifier: a system of function block-based operations for creating and manipulating meshes and mesh attributes at arbitrary complexity, with which we created six wells and distributed plaques on them. To define the surface properties of the resulting geometry, we applied materials to it, which we programmed via Blender's Shader Nodes system in a similar manner to Geometry Nodes. Lighting and camera positioning were then randomised via a mix of world shader nodes and Blender API calls. Our procedural pipeline is reproducible by relying on seed values for randomization. The only manually modelled objects in our scene were the desk in the background and the six-well  multi-titre plate, both of whose materials contained randomised details, such as variations in brightness and contrast and marker stains. After each image was rendered, it was passed through Blender's compositing pipeline, which itself is a node-based system. We rendered the background---consisting of the desk and the plate, the wells (here the actual content of the well, i.e. the stained adhesion cultured cell monolayer is meant), and the plaques in separate render layers. In the compositing step, we merged these layers together, applied post-processing effects and utilised custom render passes (called \textit{Shader AOVs} in Blender---\textit{Arbitrary Output Variables}) to export not only the enhanced base image, but also two segmentation masks: for wells, and for plaques, respectively.

\subsection*{Blender Geometry Nodes}
Using \textit{Geometry Nodes}, we first generated an array of 3x2 flat well disks.
Next, we randomly deleted the content of individual wells to prevent the model from learning a 3x2 pattern of wells.
On each of the remaining wells we then stored randomised number attributes, which were propagated downstream for further use in the procedural dye material.
We then randomly distributed plaques on each well, with a randomised density each. The plaques were modelled as tiny flat cylinders, with randomised size and Z scale to prevent overlapping geometry (known as \textit{Z fighting}). Each plaque stores its index value, as well as random values for use in the plaque materials.
Lastly, we applied the dye material to the well disks, and the plaque material to the distributed plaques.

\subsection*{Blender Materials}

The material properties and patterns were randomised through the various mesh attributes we passed on from our Geometry Nodes modifier. Both the dye material and the plaque material utilise a number of different Perlin noise \cite{perlinImageSynthesizer1985} textures and gradient calculations to resemble the patterns found in the real photographs (see \ref{fig:fig1}c-f).
For the wells, these included dark elongated gradients in each well centre, subtle high-frequency details from the drying process, as well as fully transparent areas, mainly close to the dye rims.
The dye colour hues were randomised within a colour range that we observed in the real photographs.
We achieved a semi-translucent dye effect by adding transmission to the material, and slightly reducing the alpha transparency.
In the dye material we also passed the well indices on to an AOV output, which created its own custom render pass and served as input for the well segmentation mask during the compositing step.
For the plaque material, we constructed a gradient from the plaque centre to the rim and mixed it with high-frequency Perlin noise textures to introduce irregularities.
We also passed the plaque indices---captured by the Geometry Nodes modifier---on to a shader AOV, serving as plaque segmentation masks. We could not introduce alpha transparency for the plaques themselves at this stage yet, as this would have affected the AOV output and thus the segmentation data. Instead, the plaques were rendered fully opaque over a transparent background in their own render pass, with the base colour gradient ranging from black at the rim to full white at the centre.

\subsection*{Blender Compositing}
The rendered image provided three render layers: one for the background, wells and plaques, respectively. The well and plaque layers also included our custom AOV passes for well and plaque indices. 
In this final step, we took the plaque's base colour output and interpreted it as an inverted alpha mask for the well layer. This caused the well layer to exhibit transparent holes where the plaques were supposed to be.
We then overlaid the background image with this manipulated well layer to achieve the composited image.
This composite was then run through a number of randomised filters and effects to increase the feature space of the dataset and to increase photorealism. Notably, these operations were: brightness, white balance, clarity, motion blur, low-frequency noise that acted as diffuse shadows, chromatic aberration, bloom, sharpening, and sensor noise to achieve the final image.
The well indices were exported as a segmentation mask and remapped from their original integer range to 0-1 for storage as 8-bit TIFF file. The same procedure was applied to the plaque indices. However, due to the higher number of plaques, we resorted to using 16-bit TIFF masks and as such divided the plaque index render pass by $2^{16}$ instead of $2^8$.

\subsection*{Generated Image Post-processing}
During segmentation mask rendering, Blender applies an anti-aliasing filter that introduces isolated boundary pixels with values differing from both the background and the neighbouring object. To restore binary consistency, we post-processed all masks using an eight-connected neighbourhood. Each anomalous pixel was reassigned to the most common value among its eight neighbouring pixels, thus removing isolated boundary artefacts by merging them with the adjacent object or the background.

\subsection*{Image Naturalness}
The naturalness factor $N_f$ is computed from the cumulative gradient and Laplacian distributions (CGD, CLD) of an image, as described in\cite{abgaryan2025regularized}. CGD and CLD are estimated from normalised discrete histograms of pixel-wise intensity gradients and Laplacians computed via finite differences. Both distributions are modelled as hyper-Laplace functions parametrised by scalars $T_1$ and $T_2$. These scalars were estimated from over 20k natural-scene photographs in\cite{gong2014image} to be $T_1^{\text{pr}}=0.38$ and $T_2^{\text{pr}}=0.14$. For a test image, $T_1$ and $T_2$ are obtained by fitting the CGD and CLD, and the naturalness factor is defined as
\[
N_f = (1-\theta)\frac{T_1}{T_1^{\text{pr}}} + \theta\frac{T_2}{T_2^{\text{pr}}},
\]
with $\theta=0.5$. Values of $N_f$ close to unity indicate gradient and Laplacian statistics consistent with natural-scene images, without requiring a reference image. RGB image channels were treated as individual 8-bit grayscale images for this computation, and the resulting naturalness factors were averaged.

\subsection*{Difference between Rendered and Original Image Naturalness Factor}

We evaluated the realism of our procedural pipeline by manually fine-tuning its parameters to resemble a given real image (Fig.~\ref{fig:fig1}c) as closely as possible. As such, the fine-tuned synthetic image (Fig.~\ref{fig:fig1}d) exists inside the feature space of all possible synthetic images, and at the same time allowed us to perform a direct quality comparison to the real image. The only manual change to the pipeline consisted in imported segmentation masks that resemble the plaques.
We ablated the pipeline by turning off features associated with realism, like post-processing effects or advanced material properties like transmission or high-frequency details to create synthetic versions of lower perceived realism (Fig.~\ref{fig:fig1}e-f).
We define $dN_f$ as the distance between the naturalness factor $N_{fr}$ of a real image and its synthetic counterpart, $N_{fs}$:
$$
dN_f=\left|N_{fr}-N_{fs}\right|
$$
The $dN_f$ values for the different synthetic qualities can be found in Tab. \ref{tab:metrics-realism}.

\subsection*{Fréchet Inception Distance on Image Pairs}
The Fréchet Inception Distance (FID) \cite{fid} computes the perceptual distance between two sets of images. It is typically used to assess the quality of generative image networks such as diffusion models or generative adversarial networks (GANs). It can also be used, however, to compare the quality of synthetically generated datasets, such as ours, to a ground truth set of real images. The FID is computed as the Wasserstein-2 distance between two Gaussian distributions that each model the deepest layer of an InceptionV3 model \cite{inception-v3} presented with the image set. The closer and more similar the two distributions are, the smaller the FID, and the more perceptionally similar the datasets.

Our use case required us to find a metric that could compare both pairs of images as well as pairs of datasets, since our procedural pipeline was fine-tuned on individual images first (see Fig.~\ref{fig:fig1}c-f) and then generalised to generate large datasets later. Since FID uses mean and covariance estimates from the Inception model's feature space, it typically requires large sample sizes. We stabilised the computation by splitting each synthetic image into patches of 3x2 with 15\% overlap, thus increasing the number of samples presented to the Inception network. The Gaussian distribution from the real dataset remained identical across all FID calculations; this isolated the procedural pipeline as the only variable, making the resulting scores comparable across our High, Medium, and Low realism images.

\subsection*{Dataset Diversity Metrics}
To assess image variability across the datasets, we calculated two grayscale-based descriptors - brightness variability and spatial Shannon entropy. These metrics capture distinct aspects of image heterogeneity, heterogeneity, including variation in pixel intensities and spatial uncertainty of image structures.

Brightness variability was assessed by calculating the variance of grayscale intensity values at each pixel coordinate (i,j) across all images within a dataset. This resulted in a pixel-wise variability map describing the local intensity fluctuations throughout the dataset. The mean value of this map was subsequently calculated to obtain a single measure of brightness variability for each dataset.

The spatial diversity within each dataset was quantified using a pixel-wise Shannon entropy map. This approach calculates the statistical distribution of the intensity values at each coordinate \textit{(i,j)} across all samples in the dataset. These distributions were used to compute localised entropy values, mapping spatial uncertainty in an image representing the full dataset. Finally, the spatial map was averaged to yield a single global metric representing overall structural diversity.

\subsection*{Model Evaluation Metrics} \label{subsec: metrics}
The evaluation was performed by calculating and comparing Average Precision (AP) and Intersection over Union over recalled objects (\(IoU_R\)), with both metrics averaged over all images.

Intersection over Union as defined in \cite{iou}, can be expressed with the following formula:
\begin{equation}
    IoU_R(\tau) = \sum_{o_i,\hat{o}_i\forall{i}\in(O\cap{\hat{O}})_\tau} \frac{IoU(o_i,\hat{o}_i)}{|O|} ,
\end{equation}
where \(O\) is the set of all true objects and \(\hat{O}\) is the set of predicted objects. The Intersection of those two sets is denoted by \((O\cap{\hat{O}})_\tau\) and represents the objects correctly identified with a given threshold \(\tau\in(0,1)\). The sum of False Negatives and True Positives are represented here as \(|O|\). The \(IoU_R\) metric gives a score of each correctly recalled object, which is normalised by the number of true objects.

Average Precision is defined as:
\begin{equation}
    AP(\tau) = \frac{TP_\tau}{TP_\tau + FN_\tau + FP_\tau},
\end{equation}
where \(\hat{o}\in{TP}\), if it's \(IoU\) is greater than the threshold \(\tau\). False negatives (FN) are all objects that exist, but were not predicted. Similarly, false positives (FP) are the predicted objects that don't exist in the ground truth set.

During the camera-based experiment, the results were evaluated by calculating the accuracy of plaque quantification averaged over the wells.

\subsection*{Training Loss}
The loss in the benchmark method StarDist is calculated based on the pixel-wise object probabilities and distances of the prediction \((\hat{p},  \hat{d_k})\) and their ground truths \((p, d_k)\) \cite{Stardist18}:
\begin{equation}
    L(p, \hat{p}, d, \hat{d}) = L_{obj}(p,\hat{p}) + \lambda_d L_{dist}(p, \hat{p}, d, \hat{d}).
\end{equation}

The loss \(L_{obj}\) denotes the difference between the predicted boundary distance probability map and the ground truth boundary distance probability map. It is calculated with a binary cross-entropy:
\begin{equation}
L_{obj}(p, \hat{p}) = -p \log\hat{p} - (1-p) \log(1-\hat{p}).
\end{equation}

The loss \(L_{dist}\) measures the mean absolute error of the pixel-wise errors, which are weighted by the ground truth probabilities:

\begin{equation}
L_{dist}(p, \hat{p}, d, \hat{d}) = p \cdot \mathbf{1}{_{p>0}} \cdot \frac{1}{n}\sum_k|d_k - \hat{d}_k| + \lambda_{reg} \cdot \mathbf{1}_{p=0}\cdot \frac{1}{n}\sum_k|\hat{d}_k|,
\end{equation}
where \(\lambda\) is a regularization term that is active only on true background pixels.

Measuring the loss in this specific way enhances the accuracy for points closer to the centre of an object. Afterwards, a non-maximum suppression (NMS) is performed. The goal of NMS is to exclude polygons with pixels below a certain threshold. The single shot approach from \cite{de2025single} used here applies Eqn. (3) on two parallel branches along with a custom regularisation dependent on co-location of objects detected in both branches. For further details see \cite{de2025single}.

\subsection*{SynthClaw Agentic Skill}
The \textit{synthclaw} package is designed as a modular capability system. Headless Blender rendering and analysis tools are exposed to artificial intelligence agents. The agent-facing tool interface is decoupled from the underlying graphics engine. Direct scene, material, and compositing manipulations are handled exclusively by this engine within the Blender process.

An integration, routing, and instruction framework is established by Markdown files. Skill metadata and descriptions are registered via YAML frontmatter. Automatic matching and loading by the agent framework are enabled by metadata. Execution guardrails are enforced within the document body. Furthermore, detailed operational schemas for input and output parameters are defined. The computational implementation is structured utilising Python files. 

This implementation is divided into two distinct operational layers. First, on the host side, parameter validation and timeout configurations are managed by host-side wrappers. Output directories are systematically maintained. Headless Blender instances are launched as subprocesses by these wrapper modules. Second, on the Blender side, execution scripts are run directly within the embedded Python interpreter of Blender. Compositor paths are dynamically resolved. Procedural value nodes are adjusted for scene generation. Rendering engines, such as Cycles or EEVEE, are subsequently triggered. Finally, dataset-wide diversity and realism metrics are calculated by these internal scripts.

\bibliographystyle{unsrt} 
\bibliography{refs}

\end{document}